# Coal Mine Safety Alert System: Refining BP Neural Network with Genetic Algorithm Optimization

Jiabin Luo[1(✉)] and Hanzhe Pan[2]

[1] Northeastern University at Qinhuangdao, Qinhuangdao 066004, China
[2] Northeastern University at Qinhuangdao, Qinhuangdao 066004, China
lncs@springer.com

**Abstract.** In response to the persistent safety challenges within coal mines, this study proposes a novel approach integrating a three-layer feedforward backpropagation artificial neural network with a genetic algorithm (GA-BP) for establishing a safety early warning system. Focused on a coal mine in Shandong, China, the model's effectiveness is evaluated using relevant data for training and analysis. Results indicate the superiority of the GA-BP model over traditional BP neural networks, offering enhanced capability for identifying potential safety risks promptly. This advancement enables coal mine management to implement timely interventions, ensuring the safety of miners. The findings present valuable insights for engineering applications in similar contexts.

**Keywords:** Coal mine safety, Genetic algorithm-backpropagation neural network (GA-BP), Safety early warning model, Early warning indicators.

## 1 Introduction

In recent years, coal demand in China has been growing at an average annual rate of 5%, and coal production hit a record high of of 4.5 billion tons in 2022[1]. The growth in demand for coal has driven the rapid development of the coal industry, but it has also brought about coal accidents and disasters. In the three years from 2020 to 2022, 122, 91, and 168 coal mine accidents occurred, with 225, 178, and 245 deaths, respectively[2]. In the first half of 2023, a total of 34 coal mine accidents occurred nationwide, resulting in 106 deaths[3]. It occurred on February 22,2023, the Inner Mongolia Xinjiang coal mine disaster, the death toll of 53 people, the accident since the founding of New China, China's largest accident in open-cast coal mines.

China has one of the highest rates of coal mine accidents and casualties globally, underscoring the critical need for a robust and efficient early warning model to address the severe issue of coal mine safety.

Scholars have adopted a variety of methods to study the coal mine early warning model. For example, Guo Deyong[4] applied hierarchical analysis and topology theory to establish an early warning model of coal and gas protrusion; Su Shuxian and Ouyang Mingsan[5] applied rough set and capsule neural network to construct a mine ventilation management model; Li Wenfeng and Bai Hui[6] proposed an analysis of the application of the Box-Jenkins model based on the detection of the roof safety of coal mines; Chen Hanzhang[7] proposed the design of the safety warning system for the working face of a mine based on the safety evaluation system; Ding Rijia[8] applied hierarchical analysis method, entropy value method and topology theory to construct a mine ecological early warning model for mine ecological risk level evaluation; Zhu Yana[9] studied the safety behavior evaluation and early warning of coal mine employees; Zhao Yanchao[10] researched on the application of big data based visualization predictive analytics engine in monitoring and early warning of water damage in coal mines; Qiao Wei et al[11] developed an intelligent early warning platform of big data on water damage in coal mines to realize real-time monitoring.

The research has contributed to coal mine safety but has limitations, as it considers only a few factors affecting safety, leading to potential incomplete early warnings, misjudgments, and false alarms.

Coal mine safety, being a complex nonlinear system, is challenging to address with traditional mathematical models. The BP neural network, with its strengths in automatic weight and threshold adjustment and handling large data sets and noise, is advantageous but prone to local optima and sensitivity to initial parameters. To mitigate these issues, a genetic algorithm is employed to optimize the model, enhancing its accuracy and generalization capabilities.

## 2    Related Works

The GA-BP neural network model performs nonlinear mapping by adjusting neural connections and is optimized by a genetic algorithm through selection, crossover, and mutation, enhancing predictive performance. It combines the strengths of both the BP neural network and genetic algorithm, effectively addressing complex coal mine safety issues and preventing local optimization pitfalls.

The BP neural network[12] is an artificial neural network that uses a back-propagation algorithm to forward propagate input signals through hidden and output layers, then optimizes network performance by adjusting weights and biases based on calculated error. It features input, hidden, and output layers, with neurons transforming inputs to outputs via activation functions, as depicted in Figure 1.

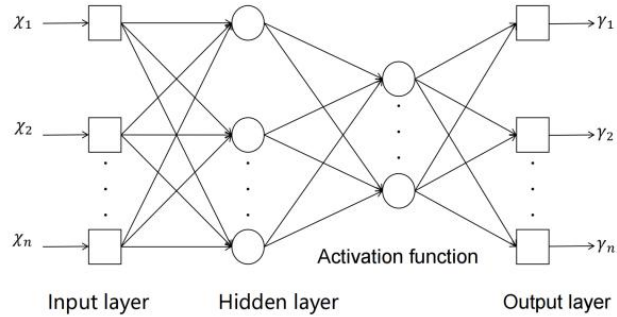

Figure 1. BP Neural Network Model.

The BP neural network comprises two main processes: forward transmission, which involves signal propagation, and backward transmission, which entails error propagation. During training, the network adjusts weights and biases iteratively by computing and back-propagating errors, aiming to minimize the difference between output and actual values. Through numerous iterations, the BP neural network acquires a robust understanding of input data features, demonstrating remarkable nonlinear fitting capabilities. Consequently, it finds applications in prediction, classification, regression, and pattern recognition tasks.

The main factors affecting coal mine safety are divided into four categories: unsafe behavior of personnel, unsafe state of objects, unsafe factors in the working environment, and unsafe organization and management, Table 1 shows the factors affecting coal mine safety indicators.

**Table 1.** The Safety Factors Affecting Coal Mines.

| Level 1 Indicators | Notation | Level 2 Indicators |
|---|---|---|
| Unsafe Behavior of Personnel | $X_{11}$ | **Security Inspections** |
| | $X_{12}$ | **Training Status** |
| | $X_{13}$ | **Technical Staff Capacity** |
| | $X_{14}$ | **Years of Experience** |
| | $X_{15}$ | **Educational attainment** |
| Unsafe State of Objects | $X_{21}$ | **Equipment Mechanization Level** |
| | $X_{22}$ | **Equipment in Good Condition** |
| | $X_{23}$ | **Firefighting Equipment Integrity Rate** |
| | $X_{24}$ | **Level of Automation of Safety Monitoring Equipment** |
| Unsafe Factors in the Working Environment | $X_{31}$ | **Coal Dust Prevention and Control** |
| | $X_{32}$ | **The Situation of Roof Prevention and Control** |
| | $X_{33}$ | **Status of Gas Prevention and Control** |
| | $X_{34}$ | **Fire Prevention and Control** |
| | $X_{35}$ | **Flood Prevention and Control** |
| Unsafe Organization and Management | $X_{41}$ | **Hidden Danger Inspection Pass Rate** |
| | $X_{42}$ | **Implementation of the Security management** |

| | $X_{43}$ | Security inspections |
|---|---|---|
| Unsafe Organization and Management | $X_{44}$ | Degree of commitment to security |
| | $X_{45}$ | Monthly safety training |

The early warning model employs a three-layer BP neural network structure, depicted in Figure 2.

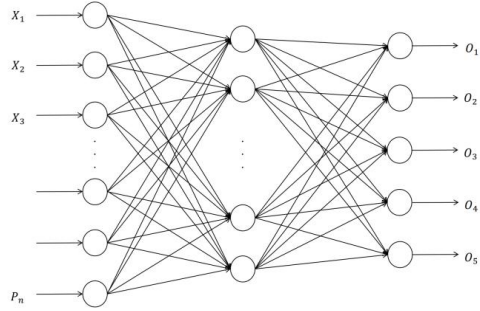

Figure 2. Design of BP Neural Network Structure for Coal Mine Safety Early Warning Assessment.

The number of neurons in the input layer, hidden layer, and output layer of the model can be determined according to the influencing factors in Table 1.

The input layer consists of 19 neurons, corresponding to the second-level index of coal mine safety.

In the output layer, there is a single neuron representing the output results of the early warning model, categorized into five levels of warning severity: high, higher, medium, lower, and low risk.

The number of neurons in the hidden layer is determined for the number of neurons in the input and output layers by the following equation.

$$n_1 = \frac{n+m}{2} + a \tag{1}$$

The number of input and output neurons falls within the integer range of [1,10].

Based on the formula, the number of neurons in the hidden layer is determined to be 11.

Genetic Algorithm is an optimization algorithm that simulates the natural evolutionary process and continuously optimizes the fitness of an individual through genetic operations such as selection, crossover, and mutation[13], and ultimately finds the optimal or near-optimal solution to the problem. It evaluates the quality of the solution based on the fitness function of the individual and can search for problems with a large solution space and find a better solution by simulating the genetic, mutation, and selection processes in nature.

Genetic algorithms (GAs) work by representing individuals, defining fitness functions, applying genetic operations (selection, crossover, and mutation), and selecting reproductive strategies to optimize individuals' fitness over time. They excel at complex optimization problems with vast solution spaces, offering robust and globally effective solutions by evaluating and enhancing individual fitness to find the best or near-best solutions.The GA-BP algorithm[14] merges genetic algorithms (GA) and backpropagation (BP) to optimize neural networks. It encodes weights and thresholds as chromosomes, evaluates them using an adaptation function, and employs selection, crossover, and mutation operations iteratively via a genetic algorithm. This process aims to enhance the neural network's performance.

First, the model's encoding mechanism involves assigning indices: input nodes as "$i$", hidden nodes as "$j$", and output nodes as "$k$". These indices correspond to four weight matrices in a BP neural network.

The weight matrix connecting the input layer to the hidden layer:

$$W = \begin{pmatrix} W_{11} & \cdots & W_{1j} \\ \vdots & \ddots & \vdots \\ W_{i1} & \cdots & W_{ij} \end{pmatrix} \tag{2}$$

Threshold matrix for the hidden layer:

$$\Upsilon = \begin{bmatrix} \gamma_1 \\ \gamma_2 \\ \cdots \\ \gamma_j \end{bmatrix} \tag{3}$$

The weight matrix from the hidden layer to the output layer:

$$V = \begin{pmatrix} V_{11} & \cdots & V_{1k} \\ \vdots & \ddots & \vdots \\ V_{j1} & \cdots & V_{jk} \end{pmatrix} \tag{4}$$

Threshold matrix for the output layer:

$$h = \begin{bmatrix} h_1 \\ h_2 \\ \cdots \\ h_k \end{bmatrix} \tag{5}$$

The above 4 matrices are expressed in the form of chromosomes. The relationship between chromosome strings and coding mappings is shown in Figure 3.

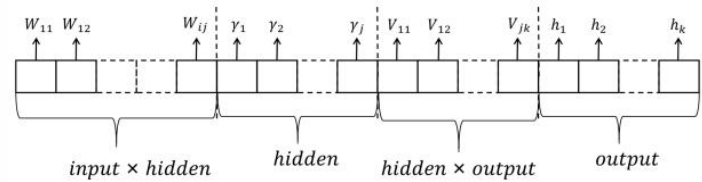

Figure 3. Chromosome and Encoding Mapping of Quantitative Traits

The chromosome length in the BP neural network, given the number of input nodes ($n$), output nodes ($m$), and hidden layer nodes ($q$), is:

$$l = (n+1) \times q + (q+1) \times m \qquad (6)$$

When substituting $n$=19, $m$=1, and $q$=11 into the formula, the length of the chromosome is given as $l$=232.

Secondly, the fitness function of the model is determined.

The set of input data is S={X11,…, X15, X21,…, X24, X31,…, X35, X41,…, X45}, the true output is $y$, after the model predicts the output to be $y^\wedge$ and $W$ is the weight matrix from the input layer to the implied layer, and $Y$ is the threshold matrix of the implied layer, and $V$ is the weight matrix from the implied layer to the output layer, and $h$ is the threshold matrix of the output layer. The forward propagation operation of the neural network is carried out and the implied layer is derived using the activation function, which yields the output of the hidden layer as $A_1$=*tansing(W1\*S+Y)*. According to the calculated $A_1$, $V$, and $h$, the model output is obtained $y^\wedge$ =$V$\*$A_1$ +$h$, the error sum of squares is taken as the objective function, and thus the fitness function is calculated. $E$=($y^\wedge$-$y$)2 as the objective function, thus calculating the fitness function $F$=$1/E$.

Finally, genetic manipulation is performed. Genetic manipulation is divided into three steps: selection, crossover, and mutation.

The selection operation in the genetic algorithm utilizes the roulette method, a strategy based on the proportion of individuals' fitnesses. Under this method, the probability of each individual ($i$) being selected is denoted as $Pi$ :

$$f_i = k / F_i \qquad (7)$$

$$P_i = \frac{f_i}{\sum_{j=1}^{N} f_i} \qquad (8)$$

Where $Fi$ represents the fitness value of individual $i$ within the population, $N$ signifies the total number of individuals in the population, and $k$ denotes the adjustment coefficient.

To introduce diversity and explore new regions in the search space, a crossover operation is employed in genetic algorithms. Given that the encoding involves real numbers, the crossover operation selects the appropriate method for real number crossover. It can better preserve small differences between genes and fully utilize the continuum structure of the search space with flexibility and broad applicability. The crossover operation between the *mth* chromosome $a_m$ and the *nth* chromosome $a_n$ at position $j$ is performed as follows:

$$\begin{cases} a_{mj} = a_{mj}(1-b) + a_{nj}b \\ a_{nj} = a_{nj}(1-b) + a_{mj}b \end{cases} \qquad (9)$$

Where $b$ is a randomly generated number between 0 and 1.

The mutation operation in genetic algorithms simulates biological variation to maintain population diversity and prevent premature convergence, thus enhancing local search capability. Common mutation algorithms include probabilistic self-adjusting mutation, uniform mutation, non-uniform mutation, Gaussian approximation mutation, and adaptive effective gene mutation. In this paper, for real number encoding, the mutation operation selects the first gene of the first individual for mutation, can help increase population diversity and improve the algorithm's exploration and search capabilities.The specific operation is as follows:

$$a_{ij} = \begin{cases} a_{ij} + f(g) * (a_{ij} - a_{\max})r \geq 0.5 \\ a_{ij} + f(g) * (a_{\min} - a_{ij})r < 0.5 \end{cases} \tag{10}$$

Where $a_{\max}$ represents the upper bound of the gene $a_{ij}$, amin is the lower bound of the gene $a_{ij}$, $f(g)$ is a function dependent on the current iteration count $g$ and the maximum number of evolutions $G_{\max}$, $r_2$ is a randomly generated number, $r$ is another randomly generated number between 0 and 1.

## 3    System Framework

To verify the feasibility and practicality of the GA-BP neural network early warning model for coal mine safety, relevant data from a coal mine in Shandong were selected for training.

The case selects data from the literature[15] for a coal mine in Shandong from 2007 to 2017 and the first two quarters of 2018.

Using min-max normalization to eliminate differences in magnitude across data as follows:

$$\begin{cases} x'_{ij} = \dfrac{x_{j\max} - x_{ij}}{x_{j\max} - x_{j\min}} \\ x'_{ij} = \dfrac{x_{ij} - x_{j\min}}{x_{j\max} - x_{j\min}} \end{cases} \tag{11}$$

Table 2 shows the data after standardization.

**Table 2.** Normalization of Samples.

| Norm | $X_{11}$ | $X_{12}$ | $X_{13}$ | $X_{14}$ | $X_{15}$ |
|------|------|------|------|------|------|
| $x_1$ | 0.112 | 0.074 | 0.210 | 0.079 | 0.107 |
| $x_2$ | 0.108 | 0.125 | 0.295 | 0.111 | 0.150 |
| ... | ... | ... | ... | ... | ... |
| $x_{46}$ | 0.906 | 1.000 | 0.769 | 1.000 | 0.481 |
| **Norm** | $X_{21}$ | $X_{22}$ | $X_{23}$ | $X_{24}$ | $X_{31}$ |
| $x_1$ | 0.131 | 0.275 | 0.329 | 0.146 | 0.072 |
| $x_2$ | 0.185 | 0.387 | 0.461 | 0.205 | 0.102 |
| ... | ... | ... | ... | ... | ... |
| $x_{46}$ | 0.900 | 0.975 | 1.000 | 1.000 | 1.000 |
| **Norm** | $X_{32}$ | $X_{33}$ | $X_{34}$ | $X_{35}$ | $X_{41}$ |
| $x_1$ | 0.098 | 0.294 | 0.210 | 0.098 | 0.206 |
| $x_2$ | 0.138 | 0.413 | 0.295 | 0.138 | 0.289 |
| ... | ... | ... | ... | ... | ... |
| $x_{46}$ | 1.000 | 1.000 | 0.970 | 0.985 | 1.000 |
| Norm | $X_{42}$ | $X_{43}$ | $X_{44}$ | $X_{45}$ | |
| $x_1$ | 0.123 | 0.330 | 0.092 | 0.750 | |
| $x_2$ | 0.171 | 0.444 | 0.130 | 0.667 | |
| ... | ... | ... | ... | ... | |
| $x_{46}$ | 1.000 | 1.000 | 1.000 | 0.500 | |

Ten sets of data are chosen from September 2008 to June 2011 as training samples, while three sets of data from December 2014 to June 2015 are selected as test samples.

The setup parameters for the GA-BP neural network model are displayed in Table 3 after selecting the appropriate parameters.

**Table 3.** Fundamental Parameter Table.

| BP Neural Network | | Genetic Algorithm | |
|---|---|---|---|
| Input Layer node | 19 | Primary | 60 |
| Hidden Layer | 1 | Numbers | 500 |
| Hidden Layer Node | 11 | Probability | 0.7 |
| Output Layer Node | 1 | Mutation | 0.05 |
| Learning Rate | 0.001 | Coding method | Real Coding |
| Performance | 0.00001 | Selection | Roulette |
| Training | trainlm | Crossover | Stochastic |

After running both the BP and GA-BP neural networks, the following results were obtained.

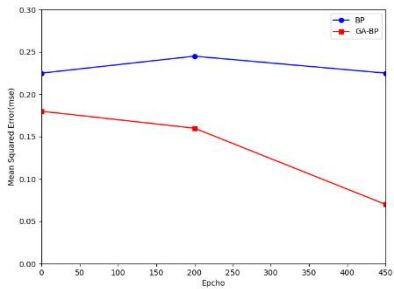

(a) Error Comparison

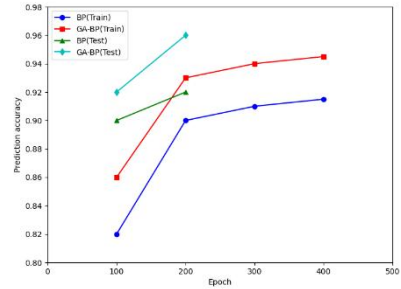

(b) Comparison of Prediction Accuracy

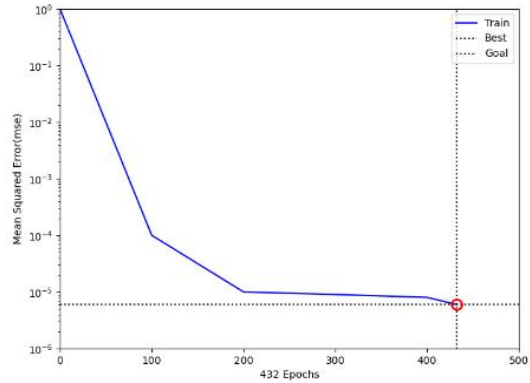

(c) GA-BP Algorithm Error Performance Curve.

Figure 4. Simulation Results

The Genetic Algorithm optimized Backpropagation (GA-BP) neural network exhibits a significant advantage over the traditional Backpropagation (BP) neural network right from the start, with initial error values below 0.20 compared to the BP network's values above this threshold. This initial edge is further enhanced by an approximate 20% reduction in error achieved through genetic algorithm optimization, leading to a substantial increase in accuracy. Analysis of Figure 4 confirms the superior performance of the GA-BP model in terms of prediction accuracy and error reduction for both training and test sets. Additionally, Figure 4 illustrates the GA-BP model's swift convergence towards the target value, highlighting its faster convergence rate and superior predictive capabilities. Collectively, these observations underscore the effectiveness and feasibility of the GA-BP neural network model for early warning systems.

While the GA-BP model demonstrates superior convergence speed and error accuracy compared to the traditional BP model, the overall operational outcomes remain unsatisfactory due to high error rates. Upon analysis, it becomes evident that the primary issue stems from the small sample size of the selected data, resulting in inadequate network training. Additionally, human factors in the data statistics of the mine contribute to the problem, along with the low quality of the acquired data.

## 4    Conclusion

An in-depth analysis of key factors affecting coal mine safety led to the development and application of a GA-BP coal mine safety early warning model using actual data from a coal mine in Shandong. Results aligned with expected objectives, indicating the model's capability to issue timely warnings and detect potential accidents, thus mitigating the occurrence of coal mine accidents and enhancing safety management. Despite these achievements, operational effectiveness remains somewhat unsatisfactory. Therefore, future research aims to enhance and optimize the early warning model to offer practical technical support for coal mine safety management.